\documentclass[paper=a4, fontsize=11pt]{scrartcl} 

\usepackage[T1]{fontenc} 
\usepackage{fourier} 
\usepackage[english]{babel} 
\usepackage{amsmath,amsfonts,amsthm} 
\usepackage{lipsum} 

\usepackage{caption}
\usepackage{subcaption}
\usepackage{graphicx}

\usepackage{hyperref}
\hypersetup{
    colorlinks=true,
    linkcolor=black,
    citecolor=black,
    urlcolor=blue,
}

\usepackage{float}
\usepackage{authblk}

\usepackage{blindtext} 

\usepackage[]{hyperref}  

\usepackage{comment}

\usepackage{sectsty} 
\allsectionsfont{\centering \normalfont\scshape} 

\usepackage{fancyhdr} 
\numberwithin{equation}{section} 
\numberwithin{figure}{section} 
\numberwithin{table}{section} 

\newcommand{\horrule}[1]{\rule{\linewidth}{#1}} 

\title{	
\normalfont \normalsize 
\horrule{0.5pt} \\[0.4cm] 
\huge Technical report: supervised training of convolutional spiking neural networks with PyTorch \\ 
\horrule{2pt} \\[0.5cm] 
}

\date{} 
\author[1, 2]{\small Romain Zimmer}
\author[2]{\small Thomas Pellegrini}
\author[1]{\small Srisht Fateh Singh}
\author[1]{\small Timoth\'ee Masquelier}

\affil[1]{\footnotesize CERCO UMR 5549, CNRS -- Universit\'e Toulouse 3, Toulouse, France}
\affil[2]{\footnotesize IRIT, Universit\'e de Toulouse, Toulouse, France}

\begin{document}
\maketitle 

\begin{abstract}

Recently, it has been shown that spiking neural networks (SNNs) can be trained efficiently, in a supervised manner, using backpropagation through time. Indeed, the most commonly used spiking neuron model, the leaky integrate-and-fire neuron, obeys a differential equation which can be approximated using discrete time steps, leading to a recurrent relation for the potential. The firing threshold causes optimization issues, but they can be overcome using a surrogate gradient. Here, we extend previous approaches in two ways. Firstly, we show that the approach can be used to train convolutional layers. Convolutions can be done in space, time (which simulates conduction delays), or both. Secondly, we include fast horizontal connections \`a la Den\`eve: when a neuron N fires, we subtract to the potentials of all the neurons with the same receptive the dot product between their weight vectors and the one of neuron N. As Den\`eve et al. showed, this is useful to represent a dynamic multidimensional analog signal in a population of spiking neurons. Here we demonstrate that, in addition, such connections also allow implementing a multidimensional send-on-delta coding scheme. We validate our approach on one speech classification benchmarks: the Google speech command dataset. We managed to reach nearly state-of-the-art accuracy (94\%) while maintaining low firing rates (about 5Hz). Our code is based on PyTorch and is available in open source at \url{http://github.com/romainzimmer/s2net}.

\end{abstract}

\newpage
\tableofcontents

\newpage
\section{Introduction}

Current Artificial Neural Networks (ANN) come from computational models of biological neurons like McCulloch-Pitts Neurons \cite{McCulloch1943} or the Perceptron \cite{Rosenblatt58theperceptron:}. Yet, they are characterized by a single, static, continuous-valued activation. On the contrary, biological neurons use discrete spikes to compute and transmit information, and spike timing, in addition to the spike rates, matters. SNNs are, thus, more biologically realistic than ANNs. Their study might help understanding how the brain encodes and processes information, and lead to new machine learning algorithms. 

SNNs are also hardware friendly and energy efficient if implemented on specialized neuromorphic hardware. These neuromorphic, non von Neumann architectures are highly connected and parallel, require low-power, and collocate memory and processing. Thus, they do not suffer from the so-called "von Neumann bottleneck" due to low bandwith between CPU and memory \cite{Backus:1978:PLV:359576.359579}. Neuromorphic architectures have also received increased attention due to the approaching end of Moore's law. Neuromorphic computers
might enable faster, more power-efficient complex calculations and on a smaller footprint than traditional von Neumann architectures. (See \cite{DBLP:journals/corr/SchumanPPBDRP17} for a survey on neuromorphic computing and neural networks in hardware).

Neuroscientists have proposed many different, more or less complex, models to describe the dynamics of spiking neurons. The Hodgkin-Huxley neuron \cite{pmid12991237} models ionic mechanisms underlying the initiation and propagation of action potentials.
More phenomenological models such as the leaky integrate-and-fire model with several variants e.g. the quadratic integrate and fire model, adaptive integrate and fire, and the exponential integrate-and-fire model have proven to be very good at predicting spike trains despite their apparent simplicity \cite{if_good_enough}. Other models such as Izhikevich's neuron model \cite{izhikevich-simple-model-of-2003} try to combine the biological plausibility of Hodgkin-Huxley-type dynamics and the computational efficiency of integrate-and-fire neurons. See \cite{if_review} for a review of the integrate and fire neuron models.

However, these models have been designed to fit experimental data and cannot be directly used to solve real life problems. 

\section{Literature review}

Various models of spiking neural networks for machine learning have alreay been proposed.

Recurrent Spiking Neural Networks (RSNNs) have been trained to generate dynamic patterns or to classify sequential data. They can have one or more populations of neurons with random or trainable connections. The computational power of recurrent spiking neural networks has been theoretically proven in \cite{MAASS2004593} and models such as liquid state machines \cite{Maass02real-timecomputing} and Long short-term memory Spiking Neural Networks (LSNNs) \cite{DBLP:journals/corr/abs-1803-09574} have been proposed.

Feed forward spiking neural networks have also been studied. \cite{10.3389/fnins.2017.00682} derives a method to convert continuous-valued deep networks to spiking neural networks for image classification. However, these models only use rate coding.

Spiking neural networks can also be trained directly using spike-timing-dependent plasticity (STDP), a local rule based on relative spike timing between neurons. This is an unsupervised training rule to extract features that can be used by an external classifier. For instance, \cite{KHERADPISHEH201856} have built a convolutional SNN trained with STDP and used a Support Vector Machine (SVM) for classification. More recently, \cite{8356226,Mozafari2019a} proposed a reward modulated version of the STDP to train a classification layer on top of the STDP network and thus, does not require any external classifier. These networks usually use latency coding with at most one spike per neuron. The label predicted by the network is given by the first spike emitted in the output layer. Backpropagation has also been adapted to this sort of coding, by computing gradients with respect to latencies~\cite{Mostafa2016a,alpha_synapses,Kheradpisheh2019}. The ``at most one spike per neuron'' limit is not an issue with static stimuli (e.g. images), yet it is not suitable for dynamic stimuli like sounds or videos.

Backpropagation through time (BPTT) \cite{bptt} cannot be used directly to train spiking neural network because of their binary activation function (see \ref{subsection:surrogate}). The same problem occurs for quantized neural networks \cite{DBLP:journals/corr/HubaraCSEB16}. However, the gradient of these functions can be approximated. For instance, \cite{DBLP:journals/corr/BengioLC13} studies various gradient estimators (e.g. straight-through estimator) for stochastic neurons and neurons with hard activation functions. Binarized networks with with performances similar to standard neural networks have been developed \cite{DBLP:journals/corr/CourbariauxB16}. They use binary weights and activations, whereas only activations are binary in this project. Yet, their encoding cannot be sparse as they use \{-1, 1\} as binary values.

These ideas can also be used to train spiking neural networks. \cite{2019arXiv190109948N} gives an overview of existing approaches and provides an introduction to surrogate gradient methods, initially proposed in ref. \cite{Bohte2000,Esser2016,Wu2017a,DBLP:journals/corr/abs-1803-09574,Shrestha2018,Zenke2018}. Moreover, \cite{DBLP:journals/corr/abs-1811-10766} proposes a Deep Continuous Local Learning (DECOLLE) capable of learning deep spatio-temporal representations from spikes by approximating gradient backpropagation using locally synthesized gradients. Thus, it can be formulated as a local synaptic plasticity rules. However, it requires a loss for each layer and these losses have to be chosen arbitrarily. Another approach has been proposed by~\cite{NIPS2018_7417}: they replaced the threshold by a gate function with narrow support, leading to a differentiable model which does not require gradient approximations.

The encoding method used in this project to represent signals with spikes (See \nameref{section:snn_sampler}) is very similar to the matching pursuit algorithm proposed by S. Mallat \cite{MP}. This algorithm adaptively decomposes a signal into a linear expansion of waveforms that are selected from a redundant dictionary of functions. Starting with the raw signal, waveforms are greedily chosen one at a time in order to maximally reduce the approximation error. At each iteration, the projection of the signal on the selected waveform is removed. The algorithm stops when the energy of the remaining signal is small enough. 
\cite{NIPS2012_4750} used a similar idea to represent efficiently a signal with the activity of a set of neurons. The potential of each neuron depends on the projection of the signal on the direction of the neuron. And each neuron compete to reconstruct the signal.
However, for this project, the goal is to classify an input signal and not to reconstruct it. Thus, the goal is to find the most interesting direction for classification and not the ones that best reconstruct the signal.
The link between send-on-delta and integrate-and-fire event-based sampling schemes has already been highlighted by \cite{quasy_iso_threshold-based_sampling}. And \cite{metric_analysis_if} proposes a mathematical metric analysis of integrate-and-fire sampling. However, they use negative spikes if the "potential" goes under the opposite of the threshold and only consider 1 dimensional input signals.

\newpage
\section{Integrate and Fire neuron models}
\subsection{Leaky Integrate and Fire (LIF)}

In the standard Leaky Integrate and Fire (LIF) model, the sub-threshold dynamics of the membrane of the $i^{th}$ neuron is described by the differential equation \cite{2019arXiv190109948N}

\begin{equation}
    \tau_{\mathrm{mem}} \frac{d U_i}{d t} = - (U_i - U_{\mathrm{rest}}) + R I_i
\end{equation}

where $U_i(t)$ is the membrane potential at time $t$, $U_{rest}$ is the resting membrane potential, $\tau_{\mathrm{mem}}$ is the membrane time constant, $I_i$ is the current injected into the neuron and R is the resistance.  When $U_i$ exceeds a threshold $B_i$, the neuron fires and $U_i$ is decreased. The $-(U_i - U_{\mathrm{rest}})$ term is the leak term that drives the potential towards $U_{\mathrm{rest}}$. 

\subsection{Non-Leaky Integrate and Fire (NLIF)}

If there is no leak, the model is called Non-Leaky Integrate and Fire (NLIF) and the corresponding differential equation is

\begin{equation}
    \tau_{\mathrm{mem}} \frac{d U_i}{d t} = R I_i
\end{equation}

Without loss of generality, we will take $R = 1$ and $U_{rest}=0$ in the following. 

\subsection{Input current}

The input current can be defined as the projection of the input spikes along the preferred direction of neuron $i$ given by $W_i$, the $i^{th}$ row of W.

\begin{equation}
    I_i = \sum_j W_{ij} S^{\mathrm{in}}_j 
\end{equation}

where $S_j(t) = \sum_k \delta(t-t_k^j)$ if neuron $j$ fires at time $t= t_1^j,t_2^j,...$.

Or, it can be a be governed by another differential equation, e.g. a leaky integration of these projections

\begin{equation*}
    \tau_{\mathrm{syn}} \frac{d I_i}{d t} = - I_i + \sum_j W_{ij} S^{\mathrm{in}}_j 
\end{equation*}

where $\tau_{\mathrm{syn}}$ is the synapse time constant.

In the former, the potential will rise instantaneously when input spikes are received whereas it will increase smoothly in the latter.

\newpage
\section{Spiking neural networks and event-based sampling}
\label{section:snn_sampler}

\subsection{Rate vs Temporal Coding}

 Standard Deep Learning (DL) is based on rate coding models that only consider the firing rate of neurons. The outputs of standard DL models are thus real-valued. However, there is evidence that precise spike timing can play an important role in the neural code \cite{Gollisch1108}, \cite{Moiseff40}, \cite{first_spikes_tactile}. Furthermore, computing with sparse binary activation can require much less computing power than traditional real-valued activation. We wanted to create a model based on precise spike timing with an efficient and sparse encoding of the information. However, there is no commonly accepted theory on how real neurons encode information with spikes. Thus, we based our approach on even-based sampling theory and explained how it can be related to spiking neural networks.
 
\subsection{Send-on-Delta}

Most of the time, the input signal is real-valued and has to be encoded as spike trains. This can be done using an event-based sampling strategy. In this work, the Send-On-Delta (SoD) sampling strategy \cite{SoD} is used.

\begin{figure}
    \centering
    \includegraphics[scale=0.7]{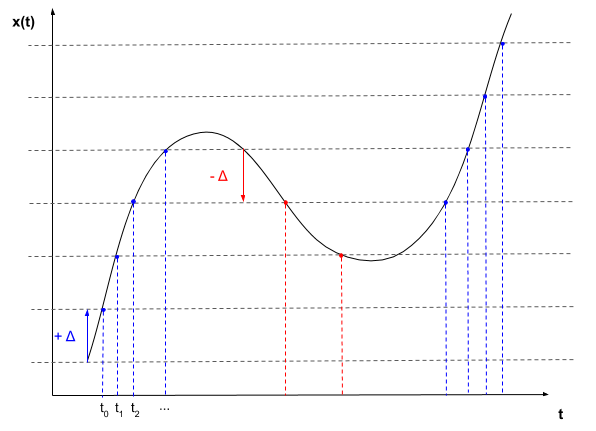}
    \caption{Send-on-delta sampling strategy. Blue dots represent sampling due to a significant increase, red dots represent sampling due to a significant decrease.}
    \label{fig:sod}
\end{figure}

The SoD strategy is a threshold-based sampling strategy. The sampling is triggered when a significant change is detected in signal $x$, i.e. the :

\begin{align}
    t_k = \mathrm{inf}_{t \geq t_{k-1}} \{t~,~|x(t) - x(t_{k-1})| \geq \Delta\} 
\end{align}

where $t_k$ is the time of the $k^{th}$ sampling event. Changes in the signal can be either an increase or a decrease (See Fig. \ref{fig:sod}).

This strategy is used by event-based cameras and is very efficient to remove temporal redundancy as sampling will only occur if the input signal changes.

\subsection{Send-on-Delta with Integrate and Fire neurons}

This encoding scheme can be achieved by two NLIF neurons with lateral connections and whose input is the derivative of the signal.

Let, 
$$I(t) = w \dot x(t) ~ \text{ and } ~ U({t_k^+}) =  0$$
with $t_k^+$ (resp. $t_k^-$) the time just after (resp. before) the $k^{th}$ spike has been emitted and $w$ a scaling factor.

Then, for the IF model we have,

\begin{equation*}
    U(t) = w(x(t)-x(t_k))
\end{equation*}

If the threshold is $B = w^2$, the next spike is emitted at $t_{k+1}^+$ such that

\begin{align*}
    t_{k+1} & = \mathrm{inf}_{t \geq t_k} \{t~,~U(t) \geq w^2\} \\
    & = \mathrm{inf}_{t \geq t_{k}} \{t~,~ \mathrm{sign}(w)(x(t)-x(t_k)) \geq |w| \}
\end{align*}

Depending on the sign of $w$, the neuron will detect an increase or a decrease of at least $|{w}|$ in the signal since the last emitted spike, provided that the potential is reset when a spike is emitted, i.e. $U_i(t_k^+) =  0$ for all $k$.

To achieve a send-on-delta sampling, two IF neurons are needed. One "ON" neuron with $w_{\mathrm{ON}}>0$ and one "OFF" neuron with $w_{\mathrm{OFF}}<0$, and their potentials have to be reset when any of them fires. Indeed, the reference value of the signal must be updated when sampling occurs. This can be done by adding lateral connections between the "ON" and "OFF" neurons. 

Let $t_k$ be the $k^{th}$ time that any of the "ON" and "OFF" neurons fires. And suppose that the "ON" neuron fires at time $t_{k+1}$. We have

\begin{align}
     U_{\mathrm{ON}}(t_{k+1}^-) & = w_{\mathrm{ON}}(x(t_{k+1}) - x(t_k))  = w_{\mathrm{ON}}^2 \\
    \implies U_{\mathrm{OFF}}(t_{k+1}^-) & = w_{\mathrm{OFF}}(x(t_{k+1}) - x(t_k)) = {w_{\mathrm{OFF}}}{w_{\mathrm{ON}}}
\end{align}

\begin{figure}
    \centering
    \includegraphics[scale=0.7]{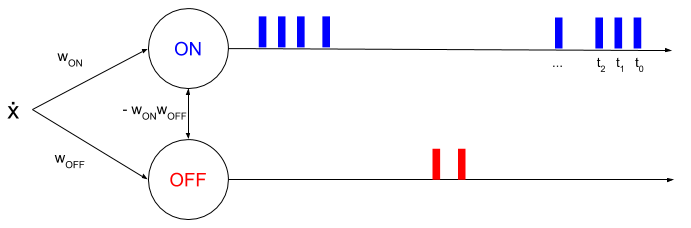}
    \caption{Architecture for SoD encoding with two IF neurons and spike train generated for the example presented in figure \ref{fig:sod}}
    \label{fig:if_sod}
\end{figure}

Applying the same reasoning for a spike emitted by the "OFF" neuron, we find that the weight of the lateral connection between the "ON" and the "OFF" neurons must be $-w_{\mathrm{OFF}}w_{\mathrm{ON}}$ in order to reset the potential of both neurons when any of them fires (see Fig. \ref{fig:if_sod}).

Note that, the reset is equivalent to an update of the input signal for each neuron. For example, if the "ON" neuron fires at time $t_{k+1}$ 

\begin{align*}
     U_{\mathrm{ON}}(t_{k+1}^-) - w_{\mathrm{ON}}^2 & = w_{\mathrm{ON}}(x(t_{k+1}) - x(t_k) - w_{\mathrm{ON}})\\
     U_{\mathrm{OFF}}(t_{k+1}^-) -w_{\mathrm{OFF}}w_{\mathrm{ON}} & = w_{\mathrm{OFF}}(x(t_{k+1}) - x(t_k) - w_{\mathrm{ON}})
\end{align*}

If an increase of at least $w_{\mathrm{ON}}$ is detected, then the reference signal is increased by $w_{\mathrm{ON}}$ for each neuron.

In this case, the deltas for the detection of an increase or a decrease are different. To have the same delta, $w_{\mathrm{OFF}}$ should be equal to $-w_{\mathrm{ON}}$.

\subsection{multi-dimensional send-on-delta}

The previous results only apply to 1-dimensional signal. For a m-dimensional signal, each dimension can be tracked independently. The number of neurons required is thus 2*m. However, tracking each dimension independently is not efficient if the coordinates of the signal are correlated. Thus, we propose a multi-dimensional generalization of send-on-delta and the corresponding network architecture, inspired by \cite{NIPS2012_4750}.

\begin{figure}
    \centering
    \includegraphics[scale=0.7]{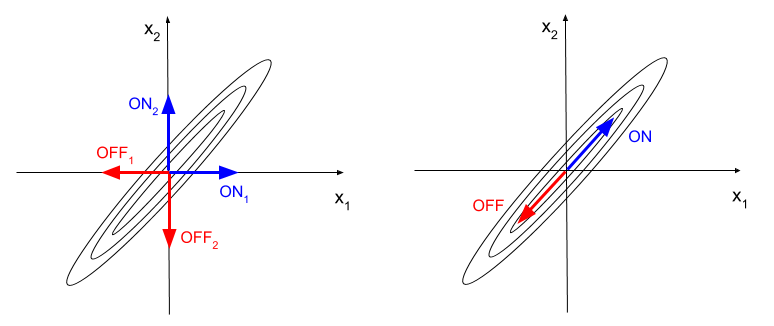}
    \caption{Left: independent tracking of each dimension, Right: tracking along a given direction}
    \label{fig:m_dim_sod}
\end{figure}

Instead of detecting changes in each dimension separately, we can detect changes in the projection of the signal along a given sampling direction. In this case, rather than simply increasing or decreasing the reference value of the signal when sampling occurs, it has now to be moved in the sampling direction (see Fig. \ref{fig:m_dim_sod}).

Let us consider a population of $n$ neurons. Each neuron $i \in [\![0,n-1]\!] $ has its own "preferred" direction $w_i$. 

After the $k^{th}$ spike emitted by any of these neurons, we have for for neuron $i$

\begin{equation*}
    U_i(t) = <w_i, x(t) - x(t_k)> 
\end{equation*}

The threshold of the $i^{th}$ neuron is set to $\|w_i\|^2$ and the weight of lateral connections between neurons $i$ and $j$ to $- <w_j, w_i>$. Just after the reset, we have

\begin{equation*}
    U_j(t_{k+1}^+) = <w_j, x(t_{k+1}) - x(t_k) - w_i> 
\end{equation*}

Thus, if neuron $i$ fires, $w_i$ is added to the signal reference value of all neurons.

Note that, if $w_i$ and $w_j$ are orthogonal the weight of lateral connections is $0$ and neurons $i$ and $j$ are independent. If $w_j$ and $w_i$ are collinear, then the potential is reset to exactly $0$. In particular, this is the case for neuron $i$.

Interestingly, the multi-dimensional generalization of SoD yields the same networks architecture as in \cite{NIPS2012_4750} for optimal spike-based representations. The only difference is that in \cite{NIPS2012_4750}, the threshold is set to $\frac{\|w_i\|^2}{2}$ instead of $\|w_i\|^2$ as they want to minimize the distance between the signal and the samples. It can be interpreted very easily in the context of event-based sampling. If sampling is associated with an update of the reference signal of $\pm \Delta$, then sampling reduces the reconstruction error as soon as the signal deviates by more than $\frac{\Delta}{2}$ from the reference.

LIF neurons can be used as well with the same architecture. The sampling scheme would be equivalent to SoD with leak. Depending on the leak, only abrupt changes would be detected.

\newpage
\section{Deep Spiking Neural Networks}

Based on \cite{2019arXiv190109948N} and the results of section \ref{section:snn_sampler}, we implemented a spiking neural network in PyTorch. Spiking Layers can be Fully-Connected or Convolutional and with or without lateral connections. The neural network can be a standard feed forward network or a pool of neurons with recurrent connections.

A PyTorch based implementation of the different layers is available at \url{http://github.com/romainzimmer/s2net}.

\subsection{LIF neurons as Recurrent Neural Networks cells}

The differential equations of LIF models can be approximated by linear recurrent equations in discrete time (See Appendix \ref{app:1}). Introducing the reset term $R_i[n]$ corresponding to lateral connections, the neuron dynamics can now be fully described by the following equations. 

\begin{align} 
    U_i[n] & = \beta (U_i[n-1]- R_i[n]) + (1-\beta) I_i[n]\\
    I_i[n] & = \sum_j W_{ij} S_j^{\mathrm{in}}[n] \\ 
    R_i[n] & = (W \cdot W^T \cdot S^{\mathrm{out}}[n])_i \\
    S_i^{\mathrm{out}}[n] & = \Theta(U_i[n] - B_i) \\
    B_i & = \| W_i\| ^2
\end{align}

where $\beta = \exp(-\frac{\Delta t}{\tau_{\mathrm{mem}}})$ and $\Theta$ is the Heaviside step function. 

Thus, LIF neurons can be modeled as a Recurrent Neural Network (RNN) cells whose state and output at time step n are given by $(U[n], I[n])$ and $S[n]$ respectively \cite{2019arXiv190109948N}.

In practice, we used a trainable threshold parameter $b_i$ for neuron $i$, such that

$$ S_i[n] = \Theta(\frac{U_i[n]}{B_i + \epsilon} - b_i) = \Theta(\frac{U_i[n]}{\| W_i\| ^2 + \epsilon} - b_i)$$ with $b_i$ initialized to 1 and $\epsilon = 10^{-8}$.
We normalize with $\| W_i\| ^2$ as the scale of the surrogate gradient is fixed.

\subsection{Surrogate gradient}
\label{subsection:surrogate}

The RNN model can be implemented with traditional Deep Learning tools. However, one major issue has to be addressed regarding the threshold activation function.

The derivative of the Heaviside step function is $0$ everywhere and is not defined in $0$. Thus, no gradient can be back-propagated through it. To solve this problem, \cite{2019arXiv190109948N} propose to approximate the derivative of the Heaviside step function.

For instance, one can approximate the Heaviside step function by a sigmoid function with a scale parameter $\sigma \geq 0$ controlling the quality of the approximation. 

Thus,

\begin{equation}
    \Theta ' \approx \mathrm{sig}_\sigma'
\end{equation}

where $\sigma \in \mathbb{R}^+$ and $\mathrm{sig}_\sigma : x \rightarrow \frac{1}{1 + e^{-\sigma x}}$

\begin{figure}
    \centering
    \includegraphics[scale=0.3]{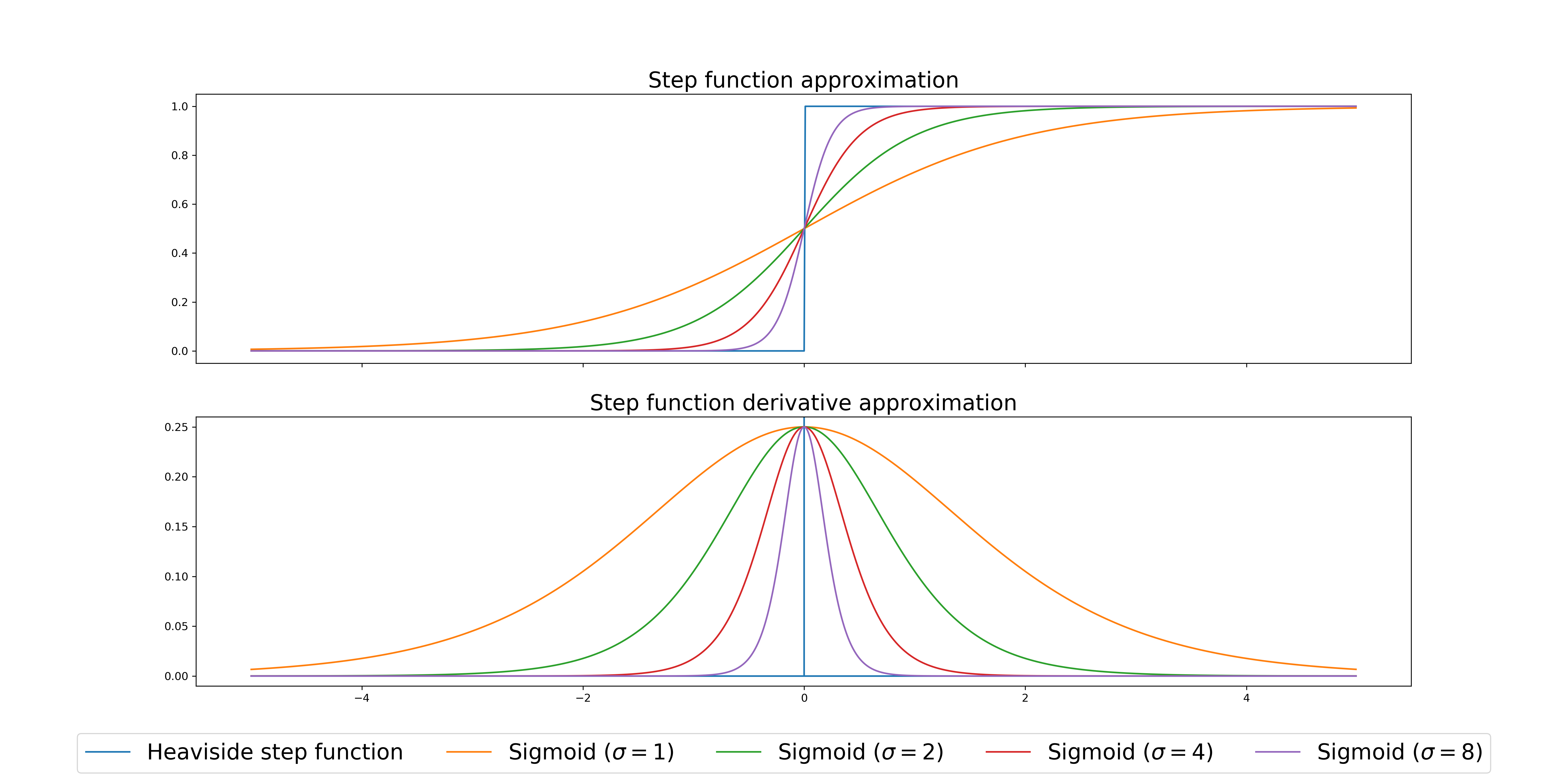}
    \caption{Approximation of the derivative of the Heaviside step function}
    \label{fig:surrogate_gradient}
\end{figure}

\subsection{Feed-forward model}

We designed a feed-forward model composed of multiple spiking layers and a readout layer. The input of a spiking layer is a spike train except for the first layer whose input is a multidimensional real-valued signal . Each layer outputs a spike train except for the readout layer that outputs real values that can be seen as a linear combination of spikes.

\subsubsection{Fully-connected spiking layer}
For the fully-connected spiking layer, the input current at each time step is a weighted sum of the input spikes emitted by the previous layer at the given moment (or a weighted sum of the input signal if it is the first layer). The state and the output of the cells are updated following the above equations.

\subsubsection{Convolutional spiking layer}
For the convolutional spiking layer, the input current at each time step is given by a 1D (0D + time),2D (1D + time) or 3D (2D + time) convolution between a kernel and the input spike train. 
Note that convolution in time can be seen as propagation delays of the input spikes. 

In this layer, lateral connections are only applied between neurons that have the same receptive field, i.e. locally between the different channels. 
For the $i^{th}$ receptive field at time step $n$, when lateral connections are used, the reset term for channel $p$ is 

\begin{equation*}
    R_{i,p}[n] = \sum_l <\tilde W_k, \tilde W_l> S_{i,l}[n-1]
\end{equation*}

Where $\tilde W_p$ is the vectorized form of $W_p$, the kernel corresponding to the $p^{th}$ channel and the sum is over the different channels.

\subsubsection{Readout layer}

For the readout layer, \cite{2019arXiv190109948N} proposed to use non-firing neurons. Thus, there is no reset nor lateral connections in this layer. For classification tasks, the dimension of the output is equal to the number of labels and the label probabilities are given by the softmax of the maximum value over time of the membrane potential of each neuron.

In practice, we have found that using time-distributed fully connected layer and taking the mean activation of this layer over time as output makes training more stable, at least with the datasets we have used. Thus, the output is the mean over time of a linear combination of input spikes.

\subsection{Recurrent Model}

In the recurrent model, a pool of neurons with recurrent connections (output is fed back to the neurons) is used instead of stacking multiple layers. The input current for this model can also be computed using convolutions. And in this case, recurrent connections are only applied locally,  i.e. between the channels for a given receptive field.

\subsection{Penalizing the number of spikes}

A simple way to penalize the number of spikes is to apply a L1 or L2 loss on the total number of spikes emitted by each layer.

However, due to the surrogate gradient, some neurons will be penalized even if they haven't emitted any spike.

As $S_k[n] \in \{0, 1\}$, the number of spikes for a given layer is

$$\# \mathrm{spikes} = \frac{1}{KN} \sum_n \sum_k S_k[n] = \frac{1}{KN} \sum_n \sum_k S_k^2[n]$$

where $K$ is the number of neurons and $N$ is the number of time steps. 

Replacing $S_k[n]$ by $S_k[n]^2$ is a simple way to ensure that the regularization will not be applied to neurons that have not emitted any spikes, i.e. for which $S_k[n]=0$. 

Indeed,

$$\frac{d S^2_k[n]}{d U_k[n]} = 2*S_k[n]*sig'_{\sigma} (U_k[n])$$

which is $0$ when $S_k[n]=0$.

\newpage
\section{Experiments}

During this project, we mainly worked with feed-forward convolutional models on the Speech Commands dataset \cite{DBLP:journals/corr/abs-1804-03209} as the goal was to compare spiking neural networks to standard deep learning models.

\subsection{Speech Commands dataset}

The Speech Commands dataset is a dataset of short audio recordings (at most 1 second, sampled at 16 kHz) of 30 different commands pronounced by different speakers for its first version and 35 for the second. All experiments were conducted on the first version of the dataset. 

\begin{table}[ht!]
\centering
\begin{tabular}{ |c|c| } 
\hline
Words (V1 and V2) & Number of Utterances \\
\hline
Bed & 2,014 \\
Bird & 2,064 \\
Cat & 2,031 \\
Dog & 2,128 \\
Down & 3,917 \\
Eight & 3,787 \\
Five & 4,052 \\
Four & 3,728 \\
Go & 3,880 \\
Happy & 2,054 \\
House & 2,113 \\
Left & 3,801 \\
Marvin & 2,100 \\
Nine & 3,934 \\
No & 3,941 \\
Off & 3,745 \\
On & 3,845 \\
One & 3,890 \\
Right & 3,778 \\
Seven & 3,998 \\
Sheila & 2,022 \\
Six & 3,860 \\ 
Stop & 3,872 \\
Three & 3,727 \\
Tree & 1,759 \\
Two & 3,880 \\
Up & 3,723 \\
Wow & 2,123 \\
Yes & 4,044 \\
Zero & 4,052 \\
\hline
Words (V2 only) & Number of Utterances \\
\hline
Backward & 1,664 \\
Forward & 1,557 \\
Follow & 1,579 \\
Learn & 1,575 \\
Visual & 1,592 \\
\hline
\end{tabular}
\caption{Number of recordings in the speech commands dataset (extracted from \cite{DBLP:journals/corr/abs-1804-03209}).}
\label{table:speech_commands_dataset}
\end{table}

The task considered is to discriminate among 12 classes:
\begin{itemize}
  \item 10 commands: "yes", "no", "up", "down", "left", "right", "on", "off", "stop", "go"
  \item unknown
  \item silence
\end{itemize}
    
The model is trained on the whole dataset. Commands that are not in the classes are labelled as unknown and silence training data are extracted from the background noise files provided with the dataset. See Table \ref{table:speech_commands_dataset}.

Authors of \cite{DBLP:journals/corr/abs-1804-03209} also provide validation and testing datasets that can be directly used to evaluate the performance of a model. 

\subsubsection{Preprocessing}

\begin{figure}
    \centering
    \includegraphics[scale=1.]{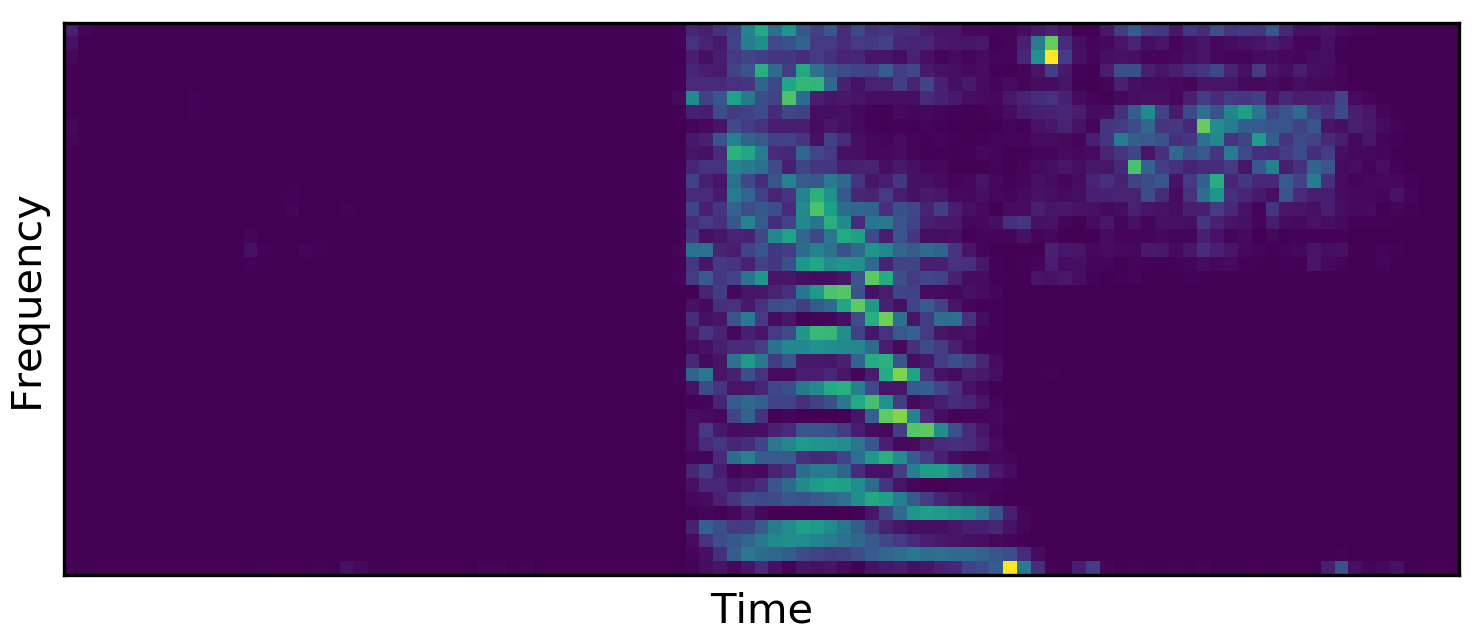}
    \caption{Example of Mel filters extracted for the word "off".}
    \label{fig:spectro_off}
\end{figure}

Log Mel filters together with their derivatives and second derivatives are extracted from raw signals using the python package LibROSA \cite{brian_mcfee_2015_18369}. For the FFT, we used a window size of 30 ms and a hop length of 10 ms, which also corresponds to the time step of the simulation $\delta_t$. These are typical values in speech processing. Then, the log of 40 Mel filter coefficients were extracted using a Mel scale between 20 Hz and 4000 Hz only as this frequency band contains most of speech signal information (see Fig \ref{fig:spectro_off}). 

Finally, the spectrograms corresponding to each derivative order are re-scaled to ensure that the signal in each frequency has a variance of 1 across time and are considered as 3 different input channels. 

\subsubsection{Architecture}

For this task, we used 3 convolutional spiking layers with the same lateral connections as in the multi-dimensional send-on-delta architecture. The readout layer is a time distributed fully connected readout layer. Each convolutional layer has a $\beta$ trainable parameter controlling the time constant of the layer, $C$ channels and one threshold parameter $b$ per channel.

Kernels are of size $H$ along the "time" axis and $W$ along the "frequency" axis. All convolutional layers have a stride of 1 and dilation factors of $d_H$ and $d_W$ along "time" and "frequency" axes respectively. See Table \ref{table:speech_commands_architecture} for details.

The scale of the surrogate gradient was set to 10.

\begin{table}[ht!]
\centering
\begin{tabular}{ |c|c c c c c| } 
\hline
Conv. layer number & $C$ & $H$ & $W$ & $d_H$ & $d_W$\\
\hline
1 & 64 & 4 & 3 & 1 & 1 \\
2 & 64 & 4 & 3 & 4 & 3 \\
3 & 64 & 4 & 3 & 16 & 9 \\
\hline
\end{tabular}
\caption{Parameters of each convolutional layer for the speech commands dataset}
\label{table:speech_commands_architecture}
\end{table}

\subsubsection{Training and evaluation}

The model was trained using the Rectified-Adam optimizer \cite{liu2019radam} with a learning rate of $10^{-3}$, for 30 epochs with 1 epoch of warm-up, a weight decay of $10^{-5}$. For each layer $l = 1,...,L$, a regularization loss $L_r(l)$ was added with a coefficient $0.1*\frac{l}{L}$ to enforce sparse activity. 

$$L_r(l) = \frac{1}{2KN} \sum_n \sum_k S_k^2[n] $$

Gradient values were clipped to $[ -5, 5 ]$, $\beta$ to $[0,1]$ and $b$ to $[0, +\infty[$  

\subsubsection{Results}

\begin{figure}
    \centering
    \includegraphics[scale=1.]{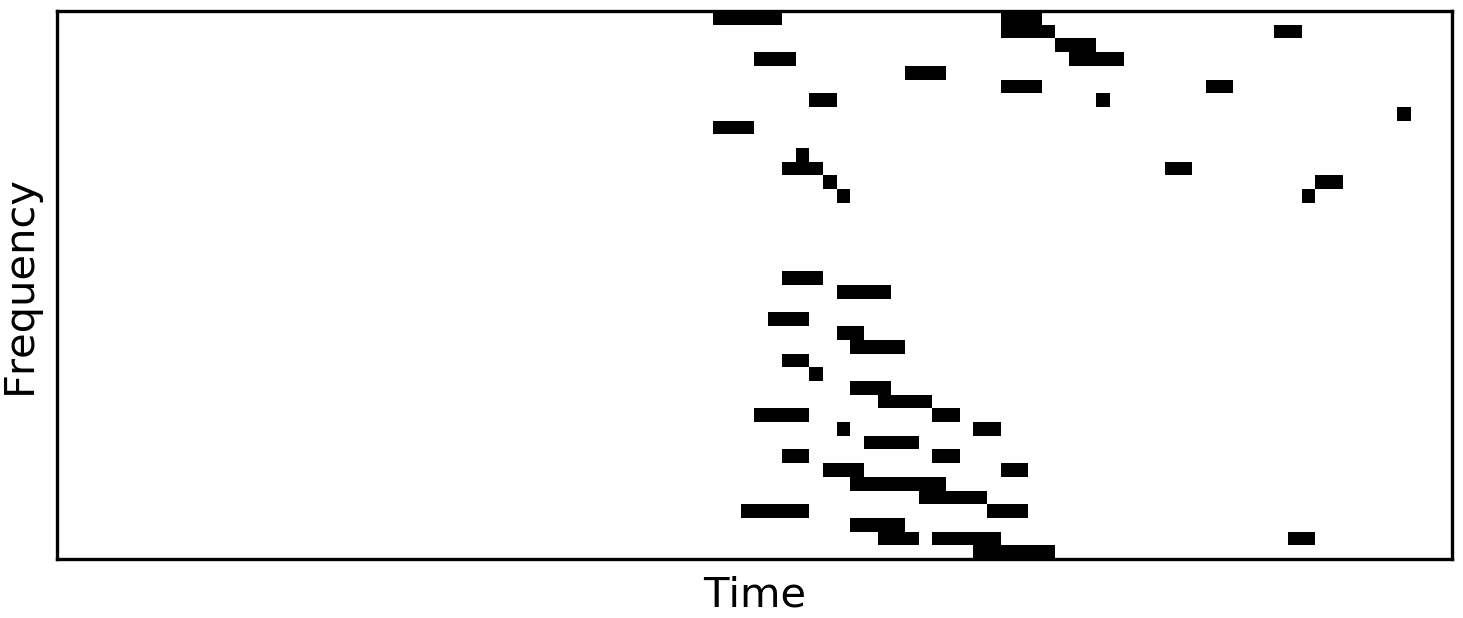}
    \caption{Example of spike train for one channel of the first layer for the word "off".}
    \label{fig:spike_train_off}
\end{figure}

This network achieves 94\% accuracy on this task with a mean firing rate of roughly 5Hz. Thus, the activation of the network is very sparse.
Moreover, the trade-off between sparsity and performance can be controlled by the regularization coefficient (see Fig. \ref{fig:spike_train_off}).

In comparison, standard deep learning models achieve an accuracy of 96-97\% \cite{journals/corr/abs-1808-08929}.

These experimental results were obtained using lateral connections. However, we managed to get similar performances without them. Thus, we would like to further explore the impact of these connections for other tasks. Furthermore, we have found that recurrence through time is not necessarily useful if the discretization time step is large and processing spikes at each time step independently is enough.

\newpage
\section{Discussion}

We proposed a generalization of the send-on-delta sampling scheme to multi dimensional signal and showed that it can be achieved by a spiking neural network with lateral connections. We designed a deep spiking neural network with binary sparse activation. This network can be trained using backpropagation through time with surrogate gradient methods and achieves comparable performance to standard deep learning models on the speech recognition task we worked on.

These results show the potential of spiking neural networks. Although PyTorch is not particularly suitable for the development of spiking neural networks, it is a very popular library in the deep learning community and we hope that this work will help to develop interest in spiking neural networks.

For future work, we would like to test our model on other tasks and especially on event data such as Dynamic Vision Sensor camera data. We would also like to continue to explore the relationships between event-based sampling theory and spiking neural networks.

\appendix
\newpage
\section{Appendix}
\subsection{Discrete time approximation} \label{app:1}

Let's consider the following differential equation \eqref{eq:E} and it's homogeneous equation \eqref{eq:H} :
\begin{align}
    \tag{E}
    \tau \frac{d u}{d t} + u & = i
    \label{eq:E} \\
    \tag{H}
    \tau \frac{d z}{d t} + z & = 0 
    \label{eq:H}
\end{align}

The solutions of \eqref{eq:E} can be found using the variation of parameters method.

The solution of \eqref{eq:H} has the following form:

\begin{equation*}
    z_K: t \rightarrow K e^{-\frac{t}{\tau}} = K z_1(t)
\end{equation*}

with $K \in \mathbb{R}$

Let's consider a solution of \eqref{eq:E} of the form

\begin{equation*}
    u: t \rightarrow k(t)z_1(t)
\end{equation*}

Injecting this solution in $E$, yields the following equivalent equation
\begin{equation*}
    k' = \frac{i}{\tau z_1}
\end{equation*}

Thus,

\begin{equation*}
    u_0 : t \rightarrow  \frac{1}{\tau} \int_0^t i(s)e^{-\frac{t-s}{\tau}} ds
\end{equation*}

is a particular solution of \eqref{eq:E} and all the solutions of \eqref{eq:E} can be written as:

\begin{equation*}
    u_K: t \rightarrow Ke^{-\frac{t}{\tau}} + \frac{1}{\tau} \int_0^t i(s)e^{-\frac{t-s}{\tau}} ds 
\end{equation*}

Now, let's consider $t,h \in \mathbb{R}$.

\begin{align*}
    u_K(t+h) = e^{-\frac{h}{\tau}} u_K(t) + \frac{1}{\tau} \int_t^{t+h} i(s)e^{-\frac{t+h-s}{\tau}} ds 
\end{align*}

For sufficiently small $h$,

\begin{align*}
    u_K(t+h) & \approx e^{-\frac{h}{\tau}} u_K(t) +  \frac{i(t+h)}{\tau} \int_t^{t+h} e^{-\frac{t+h-s}{\tau}} ds \\
    & = e^{-\frac{h}{\tau}} u_K(t) + (1 - e^{-\frac{h}{\tau}}) i(t+h)
\end{align*}

In discrete time with sampling rate $\frac{1}{h}$, \eqref{eq:E} can thus be approximated by the recurrent equation:
\begin{equation*}
    u[n] = \beta u[n-1] + (1-\beta) i[n]
\end{equation*}

with $\beta = e^{-\frac{h}{\tau}}$.

\bibliographystyle{apalike}
\bibliography{biblio.bib}

\end{document}